\pdfoutput=1
\documentclass{article}
\usepackage{iclr2027_conference,times}

\usepackage{amsmath,amsfonts,bm}

\newcommand{\figleft}{{\em (Left)}}
\newcommand{\figcenter}{{\em (Center)}}
\newcommand{\figright}{{\em (Right)}}

\def\eqref#1{equation~\ref{#1}}

\def\1{\bm{1}}

\DeclareMathAlphabet{\mathsfit}{\encodingdefault}{\sfdefault}{m}{sl}
\SetMathAlphabet{\mathsfit}{bold}{\encodingdefault}{\sfdefault}{bx}{n}

\usepackage{booktabs}
\usepackage{graphicx}
\usepackage{amsmath}
\usepackage[nointegrals]{wasysym}
\usepackage{hyperref}
\hypersetup{
  colorlinks=true,
  linkcolor=orange!50!brown,
  citecolor=teal,
  urlcolor=magenta,
  filecolor=magenta,
  pdftitle={SwaFlow: A Fully Open Suite for Studying Riemannian Flow Language Models Across Training and Scale},
  pdfauthor={Daisuke Oba, Sakae Mizuki, Hiroya Takamura, Naoaki Okazaki},
}
\usepackage{url}
\usepackage{colortbl}
\usepackage{needspace}
\usepackage{wrapfig}

\newcommand{\greyrow}{\rowcolor{black!7}}

\title{SwaFlow: A Fully Open Suite for Studying\\Riemannian Flow Language Models\\Across Training and Scale}

\author{%
  Daisuke Oba\textsuperscript{1}\thanks{Correspondence to: \texttt{daisuke.oba@nlp.comp.isct.ac.jp}} \quad
  Sakae Mizuki\textsuperscript{1,2} \quad
  Hiroya Takamura\textsuperscript{2} \quad
  Naoaki Okazaki\textsuperscript{1,2,3}\\
  \textsuperscript{1}Department of Computer Science, Institute of Science Tokyo, Japan
 \\%
  \textsuperscript{2}AIRC, National Institute of Advanced Industrial Science and Technology (AIST), Japan
 \\%
  \textsuperscript{3}LLMC, National Institute of Informatics (NII), Japan
}

\iclrfinalcopy
\begin{document}

\maketitle

\begin{abstract}
Riemannian flow language models corrupt text continuously on a curved manifold, but the models of this family released so far are final checkpoints near 0.1B parameters, which do not show how these models change with scale and over training. We release SwaFlow, a fully open suite of hyperspherical flow language models from 150M to 3B parameters trained on 2T tokens of Dolma 3 Mix, with autoregressive twins on the same data. All runs share one data order, which the released code can regenerate down to the documents behind any checkpoint. We release 386 full training states, one every 40B tokens up to 2T for each run. The suite ends in a final 3B model, mid-trained with clean prefixes and weight-averaged from its saved states. With scale, held-out loss and unconditional generation improve, and models recover tokens from noise increasingly far beyond what their embedding table guarantees. Under the standard schedule, conditioning on clean context does not improve with scale, and multiple-choice accuracy stays at chance. A short clean-prefix stage at the end of pretraining supplies this conditioning cheaply at all sizes, and the gains grow with size and budget and transfer to finetuning.
\end{abstract}
\vspace{-2.2ex}
\begin{center}\small
\begin{tabular}{@{}c@{\hspace{0.5em}}l@{}}
\raisebox{-0.28em}{\includegraphics[height=1.25em]{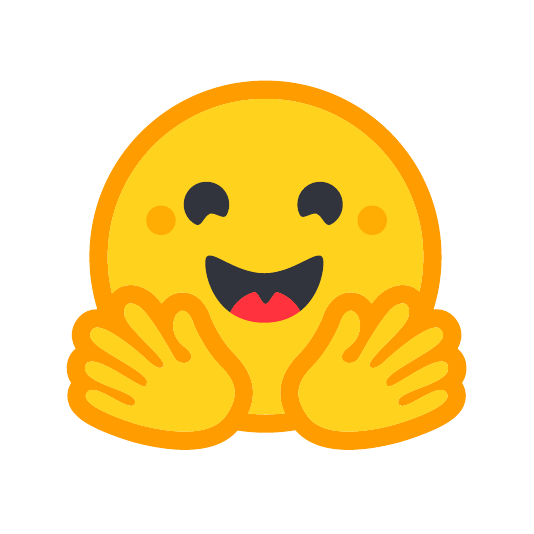}} & \href{https://huggingface.co/collections/tokyotech-llm/swaflow-suite}{\texttt{tokyotech-llm/swaflow-suite}}\\[0.15em]
\raisebox{-0.18em}{\includegraphics[height=1.0em]{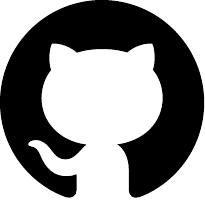}} & \href{https://github.com/swallow-llm/swaflow-suite}{\texttt{swallow-llm/swaflow-suite}}
\end{tabular}
\end{center}

\section{Introduction}\label{sec:introduction}

Riemannian flow language models corrupt text continuously on a curved manifold rather than by masking or resampling tokens \citep{sahoo2024simple,lou2024sedd,shi2024simplified,sahoo2025the}: either the statistical manifold of categorical distributions \citep{davis2024fisher,cheng2024categorical,jo2025continuous} or a hypersphere of learned token embeddings \citep{hypersphericalflows2026,sphericalflows2026}. In the hyperspherical kind, which we study, each clean embedding moves along the sphere toward random noise. A continuous ratio $\alpha$ sets how much evidence each position keeps about its token, and the geometry of the embeddings determines how informative that evidence is. Unlike a mask, which either hides a token or leaves it intact, this keeps graded evidence at every position, and what a single position reveals about its token can be computed from the embeddings and compared with what the model recovers (\S\ref{sec:shows:geometry}). Continuous language models in Euclidean space have recently been reported to rival discrete diffusion \citep{chen2026langflow,yang2026replaid} and to generate in one or a few steps \citep{fmlm2026,scalingcfm2026}; Riemannian flows share their continuous state, and whether they share these properties at scale is open.

Whether such a family is worth pursuing depends on how it behaves as it scales, and a final model cannot tell. It takes four things: several sizes trained under one recipe, their states along training, a data order shared across runs, and a control trained on the same data by next-token prediction. Open suites of autoregressive models provide much of this \citep{biderman2023pythia,groeneveld2024olmo,liu2023llm360}. For Riemannian flow language models nothing comparable exists. To our knowledge they have stayed near $0.1$B parameters while discrete diffusion and Euclidean continuous models have been trained at billions of parameters \citep{nie2025largelanguagediffusionmodels,ye2025dream,scalingcfm2026,guo2026cola}, and the models released so far are final checkpoints (Table~\ref{tab:suite}). Several come with code that reproduces individual runs, but how these models change as they grow and as they train cannot be followed from what has been released.

\begin{table}[t]
\centering\small
\caption{\textbf{Releases of Riemannian flow language models}: RDLM \citep{jo2025continuous}, Fisher-Flow \citep{davis2024fisher}, SFM \citep{cheng2024categorical}, S-FLM \citep{hypersphericalflows2026} and SFCD \citep{sphericalflows2026}. Parentheses: number of sizes.}
\label{tab:suite}
\setlength{\tabcolsep}{5pt}
\begin{tabular}{lccccc}
\toprule
 & Parameters & Max tokens & Matched AR & Released states & Data order \\
\midrule
\greyrow \multicolumn{6}{c}{\textbf{Riemannian: the statistical manifold of categorical distributions}} \\
RDLM & $92$--$110$M (2) & $0.13$T$^{\dagger}$ & -- & final only & -- \\
Fisher-Flow & $110$M (1) & $0.33$T & -- & -- & -- \\
SFM & $92$M (1) & $0.39$--$1.57$T$^{\dagger}$ & -- & -- & -- \\
\greyrow \multicolumn{6}{c}{\textbf{Riemannian: learned token embeddings on a hypersphere}} \\
S-FLM & $170$M (1) & $0.52$T$^{\dagger}$ & 1 size & final only & -- \\
SFCD & $93$M$^{\S}$ (1) & $0.52$T$^{\dagger}$ & -- & final only & -- \\
\textbf{SwaFlow} (ours) & \textbf{175M--3.7B (4)} & \textbf{4T} & \textbf{3 sizes} & \textbf{every 40B to 2T} & \textbf{recoverable} \\
\bottomrule
\end{tabular}

\vspace{2pt}
{\footnotesize $^{\dagger}$Derived: tokens as steps $\times$ batch $\times$ length. $^{\S}$Non-embedding.}
\end{table}

\begin{figure}[t]
\centering
\includegraphics[width=\textwidth]{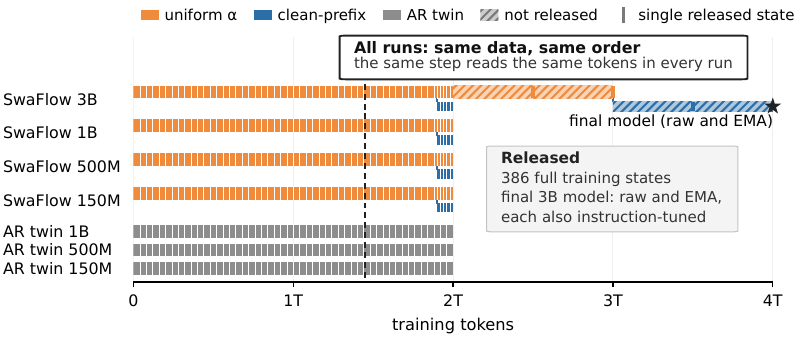}
\caption{\textbf{The suite at a glance.} All runs share one data order, so they compare step by step. Solid: released every 40B tokens (20B from 1.9T to 2T); hatched: only where marked.}
\label{fig:suite}
\end{figure}

We release \textbf{SwaFlow}, a fully open suite built to supply all four (\S\ref{sec:suite}; Fig.~\ref{fig:suite}): hyperspherical flow models of 150M, 500M, 1B and 3B parameters trained on $2$T tokens of Dolma 3 Mix \citep{olmo3}, autoregressive twins at the three smaller sizes, and full training states every $40$B tokens up to $2$T for each run, $386$ in all. The states, the final models, the data order and all code are released under permissive licenses. To our knowledge, it is the first release of Riemannian flow language models with intermediate training states or a recoverable data order (Table~\ref{tab:suite}).

The suite is built for analysis. Its data order over a public corpus is fixed by a seed and regenerated by the released code, so any checkpoint can be traced to the documents it was trained on. All runs read this order with one batch size; a given step therefore means the same tokens at every size and in the twins. Each released state is a full training state, so a run can be resumed from it; since no run decays its learning rate, a branch from any state is free to decay or not.

For users who want one checkpoint, we also release a final 3B model (\S\ref{sec:final}), mid-trained with clean prefixes at the end of training and weight-averaged from its saved states. Of the released Riemannian flow language models we can score, only SwaFlow's mid-trained models, this one included, leave chance on multiple choice, while next-token prediction on the same data stays well ahead.

We used the suite to study how a hyperspherical flow model changes with scale (\S\ref{sec:shows}). Each of the four findings below rests on something the suite provides: states along training at four sizes, branches from saved states with the continuing run as control, bases at several sizes and budgets, and a grid of saved states dense enough to average after training.

\textbf{Geometry.} The embedding table alone guarantees, in the typical case, that a token can be recovered from its position while the noise stays below a level we compute in closed form, extending a derivation for isotropic embeddings \citep{hypersphericalflows2026} to the anisotropic ones that training produces. As the model learns to recover tokens from heavier noise, growing anisotropy narrows the range the table can guarantee. Within the first fifth of training, all the noise levels the model learns from move beyond that range. Most of training therefore takes place where the geometry of a single position no longer guarantees its token, in a band that widens with scale and is where to look for the contribution of context (\S\ref{sec:shows:geometry}; Fig.~\ref{fig:band}, left).

\textbf{Conditioning.} Held-out loss and unconditional generation improve with size and budget; conditioning on clean context does not. The standard schedule, uniform $\alpha$, corrupts every position of a sequence at one level, so pretraining never shows the model a partly clean sequence (Fig.~\ref{fig:corruption}, right). A clean prompt then even makes a response harder to predict than a corrupted one, and multiple-choice accuracy stays at chance for all sizes and budgets. A clean-prefix stage at the end of pretraining, tested as a single difference against the run it branches from, removes this deficit at every size for about $1\%$ of held-out cross entropy. Afterwards a clean prompt helps and the first finetuning step comes close to an autoregressive model's. At 1B, the largest size with a twin, the model chooses a document's true continuation more often than its twin (\S\ref{sec:shows:cond}). Clean prefixes are known from Euclidean models and from task training on the sphere \citep{dieleman2022cdcd,scalingcfm2026,hypersphericalflows2026}; to our knowledge, we are the first to show what their absence costs a Riemannian flow language model and to isolate their effect in its general-corpus pretraining.

\textbf{Transfer to finetuning.} What the mid-trained bases learn in pretraining carries into finetuning. After TinyGSM finetuning \citep{liu2023tinygsm}, GSM8K accuracy \citep{cobbe2021training} with the exact sampler of the reference implementation improves with both size and pretraining budget, far ahead of a randomly initialized model; after general-purpose instruction tuning, multiple-choice accuracy grows with size (\S\ref{sec:shows:ft}).

\textbf{EMA without learning-rate decay.} Weight averaging has been proposed in place of a learning-rate decay phase for autoregressive pretraining \citep{meterez2026anytime}. It also helps a flow language model trained without decay. An exponential moving average reconstructed from the saved states improves the final model's held-out loss and multiple-choice accuracy, and the gain survives instruction tuning (\S\ref{sec:shows:ema}).

\Needspace{8\baselineskip}
\section{The SwaFlow Suite}\label{sec:suite}\label{sec:method}

\subsection{Models}\label{sec:suite:models}

\begin{figure}[t]
\centering
\includegraphics[width=\textwidth]{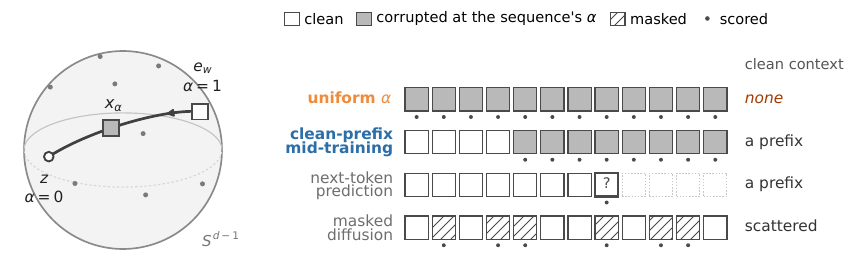}
\caption{\textbf{Corruption at one position and across a sequence.} \figleft{} A token's embedding $e_w$ moves along a great circle toward its own noise $z$ (Eq.~\ref{eq:slerp}). \figright{} One sequence per row; only uniform $\alpha$ leaves no position clean.}
\label{fig:corruption}
\end{figure}

\textbf{Objective.} SwaFlow is a hyperspherical flow language model in the sense of \citet{hypersphericalflows2026}: tokens are learned unit embeddings on $S^{d-1}$, and each clean embedding $e_w$ is corrupted along the great circle toward its own noise $z$, uniform on the hypersphere (Fig.~\ref{fig:corruption}, left),
\begin{equation}\label{eq:slerp}
x_\alpha = \frac{\sin\!\big((1-\alpha)\Theta\big)\,z + \sin(\alpha\Theta)\,e_w}{\sin\Theta}, \qquad \Theta=\arccos\langle z,e_w\rangle,
\end{equation}
with $\alpha=1$ clean and $\alpha=0$ pure noise. One $\alpha$ per sequence is drawn uniformly from a static window $[0,\alpha_{\max}]$, a schedule we call \textbf{uniform $\alpha$}, and the loss is cross entropy against the clean tokens, unweighted in $\alpha$. The window is set per size from an untruncated pilot run, at the $\alpha$ where its loss falls below $0.1$ nats, so that training is not spent on noise levels the model already solves (Table~\ref{tab:shapes}; Appendix~\ref{app:setup:train}).

\textbf{Network.} The network is a bidirectional DiT \citep{dit}, conditioned on the noise level through adaLN, with a Llama-style block \citep{grattafiori2024llama} (RMSNorm, SwiGLU, RoPE, per-head QK-norm). Table~\ref{tab:shapes} gives the four shapes.

\textbf{Autoregressive twins.} At 150M, 500M and 1B, an autoregressive model shares the network, the data, its order and the schedule, and differs only where the objective requires: causal attention, clean inputs, no noise conditioning, next-token targets.

\begin{table}[t]
\centering\footnotesize
\caption{\textbf{Model shapes}, shared by the AR twins. $N$ includes the untied embeddings.}
\label{tab:shapes}
\setlength{\tabcolsep}{4pt}
\begin{tabular}{lccccccc}
\toprule
SwaFlow & Dim & Layers & Heads & FFN & $N$ & $N$ (AR) & $\alpha_{\max}$ \\
\midrule
150M & 768  & 12 & 12 & 2048 & 0.175B & 0.161B & 0.150 \\
500M & 1280 & 24 & 10 & 3584 & 0.686B & 0.614B & 0.125 \\
1B   & 2048 & 24 & 16 & 5632 & 1.588B & 1.435B & 0.105 \\
3B   & 3072 & 28 & 24 & 8192 & 3.741B & ---    & 0.100 \\
\bottomrule
\end{tabular}
\end{table}

\subsection{Data, order and alignment across sizes}\label{sec:suite:data}

\textbf{Corpus.} The corpus is Dolma 3 Mix \citep{olmo3}, pinned to one revision, used at its published source weights and tokenized with the SmolLM2 tokenizer \citep[$V=49{,}152$;][]{smallm} into packed sequences (Appendix~\ref{app:setup:data}).

\textbf{Order.} The data order is a keyed permutation, computed on demand from a seed so that no shuffle is stored, and a run's path through it is fixed by the global batch size. All runs share both and therefore see the same tokens at any given step (Fig.~\ref{fig:suite}, dashed line), which keeps comparisons across sizes and against the twins controlled. Held-out data is set aside by the same construction and never enters a training slot.

\textbf{Tracing a checkpoint to its data.} Because the order is a function of the run's configuration and the pinned corpus revision, the released code can rebuild it, and the released order with its document index expands any step or checkpoint into the documents it consumed. The code also ships checks that the regenerated order matches what training read (Appendix~\ref{app:release:order}).

\subsection{Training recipe}\label{sec:suite:recipe}\label{sec:recover:design}

\textbf{Optimization.} After warmup the learning rate is held constant and never decayed. This choice favors analysis over final performance. Checkpoints then differ only in the data they have seen, a shorter run is a prefix of a longer one, and any state can serve as a branch point. All runs use AdamW with one batch size and a per-size learning rate (Table~\ref{tab:train}), and save a full training state, with optimizer state and dataloader position, every $20$B tokens. We implement SwaFlow on torchtitan \citep{liang2025torchtitan} and train it with FSDP2 on up to $256$ H200 GPUs (Appendix~\ref{app:setup}).

\textbf{Clean-prefix mid-training.} Under uniform $\alpha$ the model never trains on a partly clean sequence, which is what a prompt presents (Fig.~\ref{fig:corruption}, right). The last stage of the recipe supplies such sequences. Each sequence draws a split point $k$. With probability $1/2$, $k=0$ and the sequence is corrupted as under uniform $\alpha$; otherwise $k\sim\mathcal{U}\{0,\dots,1023\}$, the first $k$ positions stay clean and unscored, and the rest are corrupted at the sequence's own $\alpha$. The stage's settings were fixed in advance and never swept (Appendix~\ref{app:mid:design}). Keeping a random-length prefix clean follows \citet{dieleman2022cdcd} and \citet{scalingcfm2026}.

\textbf{The final recipe.} The final model is the 3B model, trained from scratch under uniform $\alpha$ to $3$T tokens, mid-trained with clean prefixes from $3$T to $4$T, and averaged with an exponential moving average of its weights, reconstructed after training from the saved states (\S\ref{sec:shows:ema}). Our analyses use shorter branches of the same kind, from $1.9$T to $2$T for each size with the continuing uniform run as control (Fig.~\ref{fig:suite}), and at earlier points for the 1B model.

\subsection{What we release}\label{sec:suite:release}

We release the following:
\begin{itemize}\itemsep1pt
\item \textbf{Training states.} The $386$ full states of Fig.~\ref{fig:suite}, every $40$B tokens and every $20$B where the mid-training branches are compared, listed by run in Table~\ref{tab:app:release}.
\item \textbf{Final models.} The final model, raw and averaged, each also instruction-tuned on Dolci-Instruct-SFT \citep{olmo3}.
\item \textbf{Data order.} The built order, its held-out split and a document index (Appendix~\ref{app:release:order}).
\item \textbf{Code.} Training, built on torchtitan, with the configuration of every run in this paper, including the finetuning runs whose weights are not released; the data pipeline, including the order regeneration and its checks; the evaluation harness; and the weight averaging.
\end{itemize}
Models, states and the data order are on Hugging Face (\href{https://huggingface.co/collections/tokyotech-llm/swaflow-suite}{\texttt{tokyotech-llm/swaflow-suite}}), and the code is on GitHub (\href{https://github.com/swallow-llm/swaflow-suite}{\texttt{swallow-llm/swaflow-suite}}).

Any released state can be resumed as well as evaluated, and the harness recomputes the held-out loss at every noise level from the weights, including for checkpoints we did not train (Appendix~\ref{app:release:harness}). Training saved a state every $20$B tokens, and not all of them are released (Appendix~\ref{app:release:inventory}).

\textbf{Licensing.}\label{app:release:license} Weights and our code are released under Apache-2.0, and the data order under ODC-By 1.0. The torchtitan code they build on remains under BSD-3, and parts derived from the reference implementation of \citet{hypersphericalflows2026} keep its Apache-2.0 license. The corpus, Dolma 3 Mix, is released under ODC-By 1.0, which permits redistribution with attribution, so the data behind any step can be regenerated with the released code or shared directly.

\subsection{The final model}\label{sec:final}\label{sec:post:position}

\begin{table}[b]
\centering\small
\caption{\textbf{The final models on multiple choice.} S-FLM and SFCD: released checkpoints. First row: chance. Standard errors: $.004$--$.022$ per task, $.005$ for the average. $^{\dagger}$Derived; $^{\S}$non-embedding.}
\label{tab:final}
\setlength{\tabcolsep}{3.5pt}
\resizebox{\linewidth}{!}{%
\begin{tabular}{lllllcccccc}
\toprule
Model & $N$ & $D$ & Corpus & BoolQ & OBQA & PIQA & HSwag & ARC-E & WinoG & Avg. \\
\midrule
\emph{Chance} & & & & \emph{.500} & \emph{.250} & \emph{.500} & \emph{.250} & \emph{.250} & \emph{.500} & \emph{.375} \\
\midrule
\greyrow \multicolumn{11}{c}{\textbf{Base models}} \\
S-FLM \citep{hypersphericalflows2026} & $170$M & $0.52$T$^{\dagger}$ & OWT & .482 & .252 & .492 & .263 & .269 & .507 & $.378\pm.005$ \\
SFCD \citep{sphericalflows2026} & $93$M$^{\S}$ & $0.52$T$^{\dagger}$ & OWT & .391 & .226 & .538 & .264 & .343 & .518 & $.380\pm.005$ \\
SwaFlow-3B & 3.74B & 4T & Dolma 3 & .621 & .226 & .528 & .286 & .344 & .501 & $.418\pm.005$ \\
SwaFlow-3B, EMA & 3.74B & 4T & Dolma 3 & .622 & .222 & .526 & .308 & .426 & .501 & $.434\pm.005$ \\
AR-150M & 161M & 2T & Dolma 3 & .572 & .284 & .640 & .350 & .507 & .533 & $.481\pm.005$ \\
AR-500M & 614M & 2T & Dolma 3 & .559 & .354 & .713 & .499 & .615 & .565 & $.551\pm.005$ \\
AR-1B & 1.44B & 2T & Dolma 3 & .659 & .378 & .740 & .583 & .675 & .586 & $.604\pm.005$ \\
\midrule
\greyrow \multicolumn{11}{c}{\textbf{Instruction-tuned for one epoch on Dolci-Instruct-SFT}} \\
SwaFlow-3B & 3.74B & 4T & Dolma 3 & .622 & .226 & .536 & .304 & .427 & .486 & $.433\pm.005$ \\
SwaFlow-3B, EMA & 3.74B & 4T & Dolma 3 & .622 & .238 & .556 & .312 & .457 & .485 & $.445\pm.005$ \\
AR-150M & 161M & 2T & Dolma 3 & .389 & .290 & .631 & .349 & .481 & .534 & $.446\pm.005$ \\
AR-500M & 614M & 2T & Dolma 3 & .564 & .358 & .704 & .501 & .583 & .569 & $.547\pm.005$ \\
AR-1B & 1.44B & 2T & Dolma 3 & .710 & .376 & .735 & .590 & .667 & .607 & $.614\pm.005$ \\
\bottomrule
\end{tabular}%
}
\end{table}

Table~\ref{tab:final} compares the final model with the earlier releases and the twins, before and after one epoch of instruction tuning on Dolci-Instruct-SFT (Appendix~\ref{app:post:dolci}). The earlier flow models, scored by our harness under the same rule (Appendix~\ref{app:cond:mcq}), sit at chance; the averaged 3B model leaves it, most clearly on ARC-Easy and HellaSwag, while on BoolQ it sits at the majority-answer rate. Averaging puts the model ahead of its raw weights before and after instruction tuning (\S\ref{sec:shows:ema}). As base models, the autoregressive twins score higher on average, the 150M one included.

\section{What the Suite Shows}\label{sec:shows}

Each subsection below reads one aspect of the suite along model size $N$ and training tokens $D$. Throughout, $L(\alpha)$ is the held-out cross entropy at a fixed noise level $\alpha$; the held-out data and the noise draws are fixed in advance, so every size and checkpoint is scored on identical tokens and noise (Appendix~\ref{app:release:harness}).

\subsection{Pretraining across size and budget}\label{sec:shows:pretrain}\label{sec:exp:scaling}

\begin{figure}[t]
\centering
\includegraphics[width=0.95\textwidth]{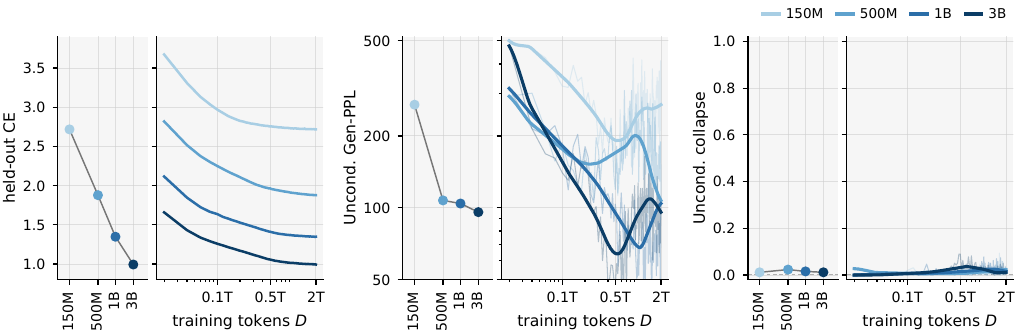}
\caption{\textbf{Unconditional pretraining scales.} Held-out cross entropy, unconditional Gen-PPL and collapse rate, against $N$ at $D=2$T (narrow left axis) and against $D$ (wide right axis).}
\label{fig:layers}
\end{figure}

Uniform $\alpha$ scales as a pretraining objective should (Fig.~\ref{fig:layers}). Averaged over a noise grid shared by all sizes, held-out cross entropy compares flow models with one another and is not a next-token perplexity. It falls with $N$ and $D$. The generative perplexity of unconditional samples, scored by an external autoregressive model as is standard \citep{sahoo2024simple,arriolablock}, falls as well, without the repetitive collapse that would make it misleading (Appendix~\ref{app:cond:gen}). The analyses below therefore concern models that train normally.

\begin{figure}[b]
\centering
\includegraphics[width=0.88\textwidth]{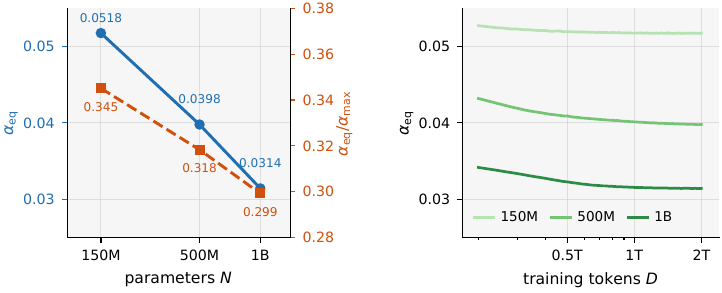}
\caption{\textbf{Equivalence point $\alpha_{\mathrm{eq}}$}, where a flow model's held-out cross entropy equals its twin's. Left: against $N$ at $D=2$T, raw and relative to the window. Right: against $D$.}
\label{fig:equivalence}
\end{figure}

\textbf{The twins as a reference.}\label{sec:exp:vs-ar} Since a flow model and its twin solve different problems, we use the twin only to fix a reference point. We take the noise level $\alpha_{\mathrm{eq}}$ at which the flow model's held-out cross entropy equals its twin's; below it, recovering a token from the corrupted sequence is harder than predicting it from a clean prefix. A larger flow model matches next-token prediction from less evidence at each position, whereas training longer barely changes where it does (Fig.~\ref{fig:equivalence}; Appendix~\ref{app:band:alphaeq}). With twins up to 1B, the trend in size rests on three points.

\subsection{Embedding geometry}\label{sec:shows:geometry}\label{sec:exp}\label{sec:exp:geo-vs-model}

In a hyperspherical flow the noise acts on learned embeddings, and the embedding table determines how much a single corrupted position reveals about its token. With states saved throughout training, the table can be tracked across training and sizes and compared with what the model solves, using two points on the held-out cross entropy $L(\alpha)$.

\begin{figure}[t]
\centering
\includegraphics[width=\textwidth]{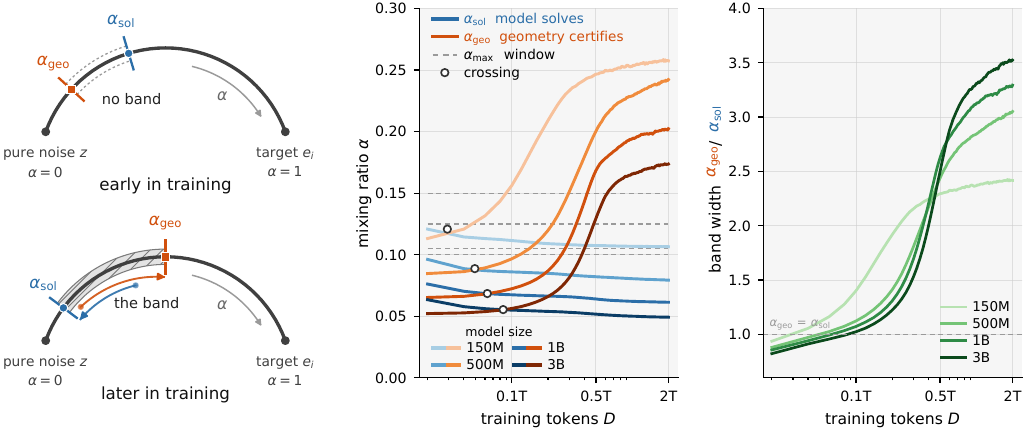}
\caption{\textbf{The information band.} \figleft{} The two boundaries before and after they cross. \figcenter{} $\alpha_{\mathrm{sol}}$ and $\alpha_{\mathrm{geo}}$ against training tokens $D$. \figright{} Band width $\alpha_{\mathrm{geo}}/\alpha_{\mathrm{sol}}$.}
\label{fig:band}
\end{figure}

\textbf{\boldmath Solvable frontier $\alpha_{\mathrm{sol}}$.} The noise level at which $L(\alpha)$ falls below $0.1$ nats. The lower it lies, the heavier the noise from which the model recovers tokens.

\textbf{\boldmath Geometric threshold $\alpha_{\mathrm{geo}}$.} The heaviest noise at which, in the typical case, a corrupted position still lies nearest its own token's embedding. As $\langle z,e_w\rangle=O(d^{-1/2})$, $x_\alpha\approx\cos\theta\,z+\sin\theta\,e_w$ with $\theta=\alpha\pi/2$: the target scores $\approx\sin\theta$ and any competitor at most $\cos\theta\,m+\sin\theta\,c$, where $m$ and $c$ are the largest cosines from $z$ to any embedding and from $e_w$ to any other. Context-free nearest-neighbor decoding thus succeeds once $\tan\theta>m/(1-c)$, and averaging $m$ over noise draws and $c$ over sampled tokens ($c_{\mathrm{mean}}$) gives a typical-case threshold set by the embedding table alone:
\begin{equation}\label{eq:alphastar}
\alpha_{\mathrm{geo}} \;=\; \tfrac{2}{\pi}\arctan\!\big(m\,/\,(1-c_{\mathrm{mean}})\big).
\end{equation}
Above $\alpha_{\mathrm{geo}}$ the embedding table guarantees recovery of the token; below it the table guarantees nothing, although decoding can still succeed there. A lower $\alpha_{\mathrm{geo}}$ means embeddings that stay separable under heavier noise, and anisotropy, which crowds neighbors together, raises it. \citet{hypersphericalflows2026} derive the threshold for isotropic embeddings, which hold only near initialization, and Eq.~\ref{eq:alphastar} has their form with one added factor, $1/(1-c_{\mathrm{mean}})$ (Appendix~\ref{app:alphastar}). Computed from the trained table, it follows the embeddings as training makes them anisotropic.

\textbf{Information band.} When $\alpha_{\mathrm{sol}}<\alpha_{\mathrm{geo}}$, the model solves noise levels at which its own embedding table no longer guarantees recovery from the position alone. We call the interval $[\alpha_{\mathrm{sol}},\alpha_{\mathrm{geo}}]$ the information band; its width $\alpha_{\mathrm{geo}}/\alpha_{\mathrm{sol}}$ measures how far the model's frontier reaches beyond that guarantee (Fig.~\ref{fig:band}, left). With one edge taken from the model's loss and the other from its embedding table, the width expresses in geometric terms how the model changes with training and scale.

\textbf{The two boundaries move in opposite directions.} Over training the model's frontier descends, more for larger models, while the threshold rises (Fig.~\ref{fig:band}, center; Table~\ref{tab:band}). The rise comes entirely from the embeddings, which lose isotropy, more so in larger models by the end of training; the noise term even drifts down (Appendix~\ref{app:band:exponents}). Embedding anisotropy is well documented \citep{mu2018allbutthetop,gao2019representation,ethayarajh2019contextual}; on the sphere, its cost can be read as the rise of $\alpha_{\mathrm{geo}}$.

\textbf{Early in training there is no band.} At first the geometry guarantees noise levels the model cannot yet solve. The two boundaries cross within the first hundred billion tokens, later for larger models, and within the first fifth of training the threshold also passes the edge of the training window (Fig.~\ref{fig:band}; Appendix~\ref{app:band:crossings}). From then on, and so for most of training, every noise level the model trains on lies beyond what per-position geometry guarantees.

\textbf{The band widens with scale.} Larger models and longer training both carry the frontier further beyond the guarantee (Fig.~\ref{fig:band}, right; Table~\ref{tab:band}). Since the models keep improving while their embeddings become less separable, the band is where the contribution of context should be measured. The released states allow a direct test at any size and checkpoint: decode each position alone and compare with the model given its full or a shuffled context.

\subsection{Conditioning on clean context}\label{sec:shows:cond}\label{sec:analysis}\label{sec:recover}

\begin{figure}[b]
\centering
\includegraphics[width=\textwidth]{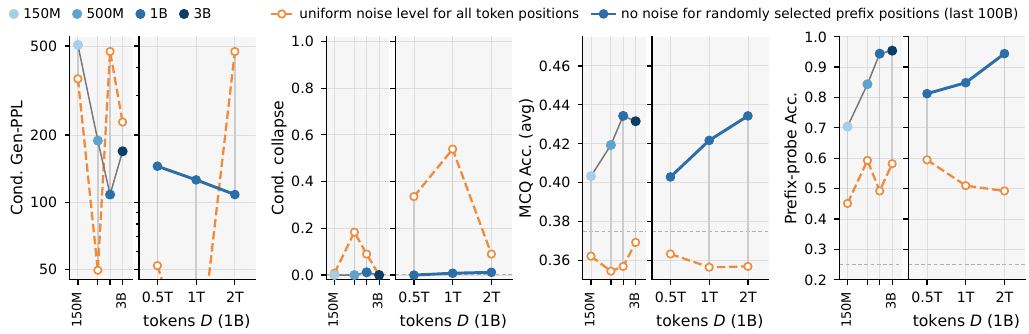}
\caption{\textbf{Conditioning under uniform $\alpha$ (open) and clean-prefix mid-training (filled).} Each pair: $N$ at $D=2$T, then 1B along $D$. Dashed: chance or real text; off-axis point: $14.9$.}
\label{fig:midtrain}
\end{figure}

The measures so far hold no position clean, as in training. Using a language model also means conditioning on clean context, and under uniform $\alpha$ every measure that requires it stays flat in $N$ and $D$ (open markers in Fig.~\ref{fig:midtrain}; Table~\ref{tab:midtrain}). Given the clean start of a held-out document, the flow models pick its true continuation among four candidates much less often than their twins (the \textbf{prefix probe}; Appendix~\ref{app:cond:probe}), and predict it worse than a unigram model that ignores the prefix. On held-out instruction data, Dolci-Instruct-SFT \citep{olmo3}, the response is harder to predict after a clean prompt than after a corrupted one. Generation from a clean prefix often collapses into repetition midway through training, more often than unconditional generation does (Appendix~\ref{app:cond:gen}). Multiple-choice accuracy also stays at chance at every size and budget, well below the twins (Appendix~\ref{app:cond:mcq}).

All of these present a sequence that is clean in one part and corrupted in the rest. Uniform $\alpha$ never produces such a sequence, since one $\alpha$ is drawn per sequence and applied to all positions (Fig.~\ref{fig:corruption}, right). Next-token prediction, in contrast, trains on nothing else, since every prediction is made from a clean prefix. A deficit on inputs the model never trains on is to be expected; the question the suite answers is whether scale removes it. Neither $N$ nor $D$ does, so we attribute the deficit to this gap in the training distribution rather than to capacity or budget.

We test the attribution by changing only the schedule, with the clean-prefix mid-training of \S\ref{sec:suite:recipe} applied as a branch late in training so that each model's own run is the control. Each model leaves its uniform run at $1.9$T and trains to $2$T under the new schedule, while the uniform run continues over the same $100$B tokens with its learning rate unchanged. The two runs then differ only in which positions are held clean, and neither sees instruction data (Appendix~\ref{app:mid:design}). For the budget axis, the 1B model is also branched to land at $0.5$ and $1.0$T. Clean prefixes are a known ingredient for prompting and for faithful samples \citep{dieleman2022cdcd,scalingcfm2026,hypersphericalflows2026}; here they serve as a controlled test of the diagnosis.

\begin{table}[t]
\centering\small
\newcommand{\betterup}[1]{\ensuremath{_{\textcolor{green!55!black}{+#1}}}}%
\newcommand{\betterdown}[1]{\ensuremath{_{\textcolor{green!55!black}{-#1}}}}%
\newcommand{\worseup}[1]{\ensuremath{_{\textcolor{red}{+#1}}}}%
\newcommand{\neutral}[1]{\ensuremath{_{\textcolor{gray}{#1^{\dagger}}}}}%
\newcommand{\dzero}[1]{\ensuremath{_{\pm#1}}}%
\caption{\textbf{Uniform $\alpha$ against clean-prefix mid-training at $D=2$T.} SFT-format rows: held-out instruction responses, prompt clean or corrupted. $\|g\|$: first-step gradient norm of instruction tuning. Subscripts: change from uniform $\alpha$, green if better and red if worse. $^{\dagger}$Gray: conditional Gen-PPL is better or worse only when read with the collapse rate.}
\label{tab:midtrain}
\setlength{\tabcolsep}{3pt}
\resizebox{\linewidth}{!}{%
\begin{tabular}{lccccccccc}
\toprule
 & \multicolumn{4}{c}{uniform $\alpha$} & & \multicolumn{4}{c}{clean-prefix mid-training} \\
\cmidrule(r){2-5}\cmidrule(l){7-10}
 & 150M & 500M & 1B & 3B & & 150M & 500M & 1B & 3B \\
\midrule
MCQ Acc. (chance $0.375$)    & 0.362 & 0.354 & 0.357 & 0.369 & & 0.403\betterup{0.041} & 0.419\betterup{0.065} & 0.434\betterup{0.077} & 0.432\betterup{0.063} \\
Prefix-probe Acc. (chance $0.25$)     & 0.451 & 0.593 & 0.492 & 0.582 & & 0.704\betterup{0.253} & 0.844\betterup{0.251} & 0.944\betterup{0.452} & 0.954\betterup{0.372} \\
Clean-prefix CE (unigram $7.53$) & 8.00 & 8.37 & 7.72 & 8.53 & & 7.16\betterdown{0.84} & 7.07\betterdown{1.30} & 6.88\betterdown{0.84} & 6.92\betterdown{1.61} \\
\midrule
Cond. collapse rate (\%)   & $0.8$ & $18.4$ & $9.0$ & $0.0$ & & 0.0\betterdown{0.8} & 0.0\betterdown{18.4} & 1.2\betterdown{7.8} & 0.0\dzero{0.0} \\
Cond. Gen-PPL (real text $9.64$) & 356 & 50 & 472 & 228 & & 505\neutral{+149} & 189\neutral{+139} & 108\neutral{-364} & 169\neutral{-59} \\
\midrule
SFT-format CE, clean prompt      & 4.010 & 3.080 & 3.151 & 2.109 & & 2.411\betterdown{1.599} & 1.652\betterdown{1.428} & 1.162\betterdown{1.989} & 0.860\betterdown{1.249} \\
SFT-format CE, noised prompt     & 2.492 & 1.695 & 1.205 & 0.892 & & 2.519\worseup{0.027} & 1.714\worseup{0.019} & 1.221\worseup{0.016} & 0.905\worseup{0.013} \\
$\|g\|$ at fine-tuning step 1     & 9.62 & 21.19 & 38.87 & 15.81 & & 1.87\betterdown{7.75} & 1.68\betterdown{19.51} & 1.56\betterdown{37.31} & 0.97\betterdown{14.84} \\
\midrule
Held-out CE (common grid)        & 2.718 & 1.879 & 1.348 & 0.994 & & 2.738\worseup{0.020} & 1.896\worseup{0.017} & 1.360\worseup{0.012} & 1.006\worseup{0.012} \\
\bottomrule
\end{tabular}%
}
\end{table}

\textbf{Conditioning appears without finetuning.}\label{sec:recover:main} Mid-training removes each of these failures at all four sizes (Fig.~\ref{fig:midtrain}; Table~\ref{tab:midtrain}). The prefix probe rises well above the uniform runs, a clean prefix now brings the loss of what follows below the unigram level, and generation from a clean prefix no longer collapses. The price is about $1\%$ of held-out cross entropy, an order of magnitude less than one epoch of instruction tuning costs the same checkpoints, and unconditional generation is unchanged (Appendix~\ref{app:mid:cost}).

\textbf{In the loss.}\label{sec:recover:mech} After mid-training a clean prompt helps predict the response at all four sizes, while the loss under a corrupted prompt barely moves (Table~\ref{tab:midtrain}); the use of clean context is gained without giving up the task the model was trained on.

\begin{figure}[t]
\centering
\includegraphics[scale=0.86]{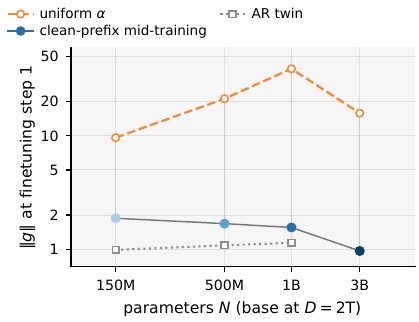}
\caption{\textbf{Gradient norm at the first finetuning step}, from bases at $D=2$T.}
\label{fig:gradnorm}
\end{figure}

\textbf{At the start of finetuning.} The first gradient of an identical instruction-tuning run shows how far a base is from the task (Fig.~\ref{fig:gradnorm}). From uniform-$\alpha$ bases it is an order of magnitude larger than from the twins; after mid-training it falls to within a factor of two of theirs, so a mid-trained flow model enters finetuning about where an autoregressive one does. Finetuning on top of mid-training accordingly adds less multiple-choice accuracy than mid-training itself (Appendix~\ref{app:mid:both}).

\textbf{Multiple choice.} Every mid-trained model leaves chance. Much of the gain is on BoolQ, a yes/no task on which the models move from the always-no rate to near the always-yes rate; BoolQ alone therefore does not show that the passage is read. The other tasks move as well, most clearly ARC-Easy at the larger sizes (Appendix~\ref{app:mid:main}).

\textbf{Along $N$ and $D$.}\label{sec:recover:daxis} The gains grow with size on both the probe and multiple choice, and since BoolQ stays flat across sizes, the growth in multiple choice comes from the other tasks. On the probe the mid-trained models overtake their twins by 1B; on multiple choice the twins stay ahead. At 1B the probe gain also grows with the budget at which mid-training is applied, while the uniform run declines (Fig.~\ref{fig:midtrain}, right of each pair). Multiple choice rises along $D$ as well, but this rise is carried by BoolQ (Appendix~\ref{app:mid:daxis}). As every branch spends $100$B tokens, the growth along $D$ reflects what the preceding pretraining provides. A clean prompt already helps at the earliest branch (Appendix~\ref{app:mid:daxis}), so the further gains come from what the model builds on that ability.

\subsection{Transfer to finetuning}\label{sec:shows:ft}\label{sec:post}\label{sec:post:tinygsm}

\textbf{Task-specific finetuning.} We test whether what a mid-trained base learned in pretraining transfers to task-specific finetuning. Each base is finetuned on TinyGSM \citep{liu2023tinygsm} for up to $65$B tokens, alongside a randomly initialized model of the same size (``none''), and scored on GSM8K \citep{cobbe2021training} $0$-shot by executing the generated program, with two samplers, \emph{exact} and \emph{top-1} (Appendix~\ref{app:post:setup}).

\begin{figure}[t]
\centering
\includegraphics[width=\textwidth]{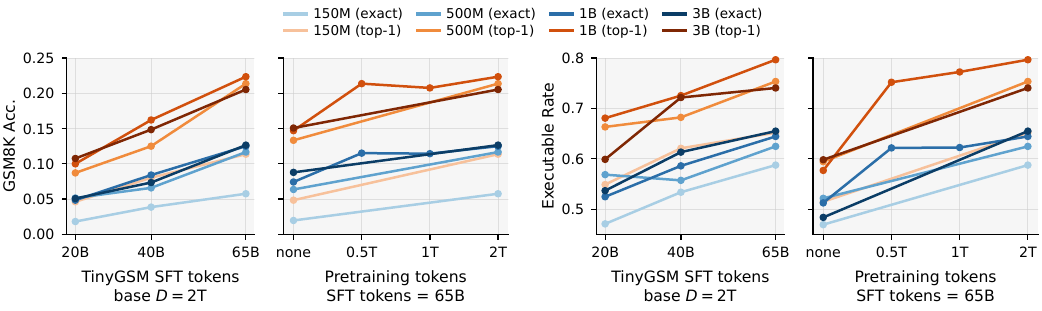}
\caption{\textbf{Task-specific finetuning on TinyGSM.} GSM8K accuracy (left pair) and executable rate (right pair), along finetuning tokens from $2$T bases and along the base's pretraining tokens (``none'': random initialization). Blue: exact sampler; orange: top-1; darker: larger.}
\label{fig:tinygsm}
\end{figure}

Pretraining transfers (Fig.~\ref{fig:tinygsm}). At every size a pretrained base finishes well ahead of a random start, and with the reference implementation's exact sampler, accuracy improves with both $N$ and $D$, with the rate of executable programs rising alongside (Appendix~\ref{app:post:axes}). Accuracy is still rising at the end of finetuning, so these values are not ceilings.

\textbf{General-purpose instruction tuning.} Every $2$T base, flow and autoregressive, is also instruction-tuned for one epoch on Dolci-Instruct-SFT (Appendix~\ref{app:post:dolci}). From mid-trained bases, multiple-choice accuracy after tuning grows with size (Table~\ref{tab:app:dolci}). Instruction tuning also lifts the uniform-$\alpha$ bases off chance, since its prompts are clean, but it costs them about ten times more held-out cross entropy than mid-training does, and they end behind the mid-trained bases at all sizes. The twins remain ahead after the same tuning.

\subsection{EMA without learning-rate decay}\label{sec:shows:ema}

Weight averaging carries over to the pretraining of a flow language model. Learning-rate schedules usually end in a decay phase, which ties a run to a horizon fixed in advance; \citet{meterez2026anytime} instead keep the schedule horizon-free and average the weights, with an exponential moving average whose decay $\tau_t=2^{-f/t}$ follows the step $t$. Their evidence comes from autoregressive language models. SwaFlow can test this for a flow language model: no run decays its learning rate, and with a full state saved every $20$B tokens, the average can be reconstructed after training without a further run (Appendix~\ref{app:mid:ema}).

On the final 3B model, the average ($f=100$) lowers held-out cross entropy from $1.006$ to $0.968$ and raises multiple-choice accuracy, and the lead survives instruction tuning (Table~\ref{tab:final}). A flow model trained without learning-rate decay can thus be improved from its saved states. The released code implements the reconstruction, which we ran on the full $20$B grid of saved states; before $1.9$T, the released states are $40$B apart (Appendix~\ref{app:release:inventory}).

\section{Related Work}\label{sec:relatedwork}

\textbf{Open model suites.} For autoregressive language models, releasing a family of models with their intermediate checkpoints, data and code has become a common way to study how behavior develops across training and scale \citep{biderman2023pythia,groeneveld2024olmo,olmo2,olmo3,liu2023llm360}. SwaFlow brings this practice to Riemannian flow language models, whose earlier releases contain only final checkpoints (Table~\ref{tab:suite}). It keeps the design in which all sizes read one data order \citep{biderman2023pythia}, and adds autoregressive twins on that order to isolate the effect of the objective.

\textbf{Riemannian flows over discrete data.} Of the two lines in this family, only the one that moves learned token embeddings on a sphere \citep{hypersphericalflows2026,sphericalflows2026} has a geometry that training shapes; on the statistical manifold of categorical distributions, each category is a fixed vertex \citep{davis2024fisher,cheng2024categorical,cheng2025alphaflow,jo2025continuous}. The $c_{\mathrm{mean}}$ and $\alpha_{\mathrm{geo}}$ of \S\ref{sec:exp} therefore exist only on the sphere. We keep the objective of \citet{hypersphericalflows2026} and its truncation of the noise range, with the bound set from a pilot run, and change the scale (Table~\ref{tab:suite}). We claim no advantage for the manifold itself; in an ablation, \citet{fmlm2026} find a Euclidean interpolant ahead of Riemannian diffusion.

\textbf{Euclidean continuous diffusion and flows.} Continuous language models outside this family corrupt in Euclidean space: learned embeddings under Gaussian noise \citep{li2022diffusion,dieleman2022cdcd,gulrajani2023plaid,chen2026langflow,yang2026replaid}, one-hot vectors \citep{fmlm2026,scalingcfm2026} or encoder and autoencoder latents \citep{elf2026,guo2026cola}. Some have been trained at one to a few billion parameters \citep{gulrajani2023plaid,scalingcfm2026,guo2026cola}, but, to our knowledge, none releases intermediate states or its data order. In this line, clean prefixes make models promptable and their samples faithful \citep{dieleman2022cdcd,scalingcfm2026}. We pin down the failure they repair and test them against a matched control (\S\ref{sec:shows:cond}). Trained with them throughout, the model of \citet{scalingcfm2026} scores above chance on some multiple-choice tasks.

\textbf{Discrete diffusion.} Masked diffusion has scaled further, with fitted laws \citep{nie2024scalingmdm} and $7$--$8$B models \citep{nie2025largelanguagediffusionmodels,ye2025dream}. Tokens left unmasked stay clean, and these models therefore train on partly clean sequences (Fig.~\ref{fig:corruption}, right).

\textbf{Scaling laws.} Scaling studies fit one held-out loss against $N$ and $D$ \citep{kaplan2020scaling,hoffmann2022training}, and continuous diffusion over text has its own fits, which place it at a large compute multiple of autoregression \citep{gulrajani2023plaid,yang2026replaid}; RePlaid also reads its loss per diffusion timestep, at one size. We instead follow held-out loss along $\alpha$, an axis that autoregressive training does not have, across $N$ and $D$: the loss can fall at every $\alpha$ while the frontier moves toward heavier noise.

\textbf{Post-training and conditioning.} Conditioning on a clean instruction is usually obtained afterwards through instruction tuning \citep{wei2022flan,ouyang2022instructgpt}, and late pretraining usually changes the data \citep{gururangan2020dontstop,ibrahim2024continual,hu2024minicpm,olmo2}, masking patterns aside \citep{bavarian2022fim,tay2023ul2}. Our clean-prefix stage instead keeps the data and changes the objective, the source of the deficit here.

\section{Conclusion}\label{sec:conclusion}

Earlier releases of Riemannian flow language models consist of final checkpoints near $0.1$B parameters; SwaFlow opens this family to study across training and scale. The suite holds hyperspherical flow models from 150M to 3B parameters trained on $2$T tokens of a public corpus, autoregressive twins at three sizes, and full training states every $40$B tokens, all on one data order that the released code can regenerate.

The analyses in this paper show what such a suite makes visible. With scale, held-out loss and unconditional generation improve, and the noise levels at which the models recover tokens move ever further past the guarantee of their embedding table. Conditioning on clean context is the exception. It does not improve with scale, and we trace the deficit to uniform $\alpha$, under which pretraining never presents a clean prompt. A short clean-prefix stage at the end of pretraining supplies it at all sizes at little cost, and its gains grow with size and budget and carry into finetuning. Weight averaging, proposed in place of learning-rate decay for autoregressive models, also improves a flow model trained without decay. The final 3B model is built from both, mid-trained with clean prefixes and averaged from its saved states.

\subsection{Discussion and Limitations}\label{sec:discussion}

Because the geometric threshold is a typical-case, sufficient condition, the information band shows where the embedding table stops guaranteeing recovery, but not what the model relies on inside it; a context-free decoder and context ablations on the released states would measure that. The evidence that mid-trained models use clean context rests on the prefix probe and on the loss of instruction responses, since much of the multiple-choice gain is on BoolQ, where it coincides with a shift toward the majority answer. The twins stop at 1B, the clean-prefix stage was set before any run and never tuned, and generation speed was not compared with the twins. A run resumed from the released states draws a fresh noise stream, as they do not carry the random-number state, and our corpus omits the suffix-array deduplication marked in its web split (Appendix~\ref{app:setup:data}).

Our diagnosis concerns drawing one noise level per sequence and does not depend on the sphere; whether other continuous language models share the deficit and respond to the remedy remains to be tested. Within this family, two questions about the remedy are open: whether clean prefixes throughout pretraining would condition better than a late stage, and how much of the remaining gap to the twins lies in the objective rather than in the training regime. The frontier $\alpha_{\mathrm{sol}}$ can be read off any model with a noise-level axis, whereas $\alpha_{\mathrm{geo}}$ needs a learned embedding table; for embeddings normalized to a fixed length under a variance-preserving Gaussian path \citep{chen2026langflow,yang2026replaid}, an analog appears derivable.

SwaFlow is a resource for research rather than a set of models for applications. It opens Riemannian flow language models to study across training and scale, and its autoregressive twins measure how far this family still has to go. As base models, the twins score higher on multiple choice than the final 3B model (\S\ref{sec:final}), and samples from the flow models remain far from natural text in generative perplexity (Fig.~\ref{fig:layers}). None of the models, the instruction-tuned ones included, has been evaluated for safety.

\section*{Acknowledgments}
This research was conducted by AIST and Institute of Science Tokyo under FRONTia, a Japanese national program led by the Ministry of Economy, Trade and Industry of Japan (METI) and the New Energy and Industrial Technology Development Organization (NEDO), aimed at developing a domestic multimodal foundation model for AI robots and physical AI.

\bibliography{iclr2027_conference,refs_added}
\bibliographystyle{iclr2027_conference}

\appendix

\section{The release in detail}\label{app:release}

\subsection{Inventory}\label{app:release:inventory}

\begin{table}[h]
\centering\small
\caption{\textbf{Released training states}, $386$ in all; spacing in parentheses. Final models are also released as weights.}
\label{tab:app:release}
\setlength{\tabcolsep}{5pt}
\begin{tabular}{lll}
\toprule
Model & Uniform $\alpha$ & Clean-prefix \\
\midrule
Flow 150M, 500M, 1B & $0.04$--$1.88$T ($40$B); $1.9$--$2$T ($20$B) & $1.92$--$2$T ($20$B) \\
Flow 3B & $0.04$--$1.88$T ($40$B); $1.9$--$2$T ($20$B); $2.5$T, $3$T & $1.92$--$2$T ($20$B); $3.5$T, $4$T \\
Autoregressive 150M, 500M, 1B & $0.04$--$2$T ($40$B) & --- \\
\bottomrule
\end{tabular}
\end{table}

\textbf{Random-number state.} The released states do not include the generator state behind the $\alpha$ and noise draws. A run resumed or branched from them restores optimizer, schedule and dataloader exactly but draws a different, identically distributed noise stream. Bitwise reproduction across hardware is not expected either.

\textbf{States not released.} Training saved a full state every $20$B tokens, and every analysis in this paper reads that grid; three kinds of state are not released. Before $1.9$T, only every other state is released, so the released states lie $40$B apart; the $20$B spacing is kept from $1.9$T to $2$T, where the mid-training branches are compared. The 1B branches mid-trained to $0.5$T and $1$T (\S\ref{sec:shows:cond}; the budget axis of \S\ref{sec:shows:ft}) are not released; the first can be re-created by branching the released $0.4$T state with the released code, while the second starts from the $0.9$T state, which is not released. Beyond $2$T, the 3B model is released at $500$B intervals rather than every $20$B; the averaged final model was reconstructed from the full $20$B grid and is released as weights.

\subsection{The harness and its \texorpdfstring{$\alpha$}{alpha} grids}\label{app:release:harness}

\textbf{Grids.} The harness scores saved states after training. Cross entropy is compared across sizes on the \emph{common grid}, eight points on $[0,0.10]$. $\alpha_{\mathrm{sol}}$ is read on the \emph{window grid}, $32$ points over each size's window, by interpolating in log cross entropy, since at 150M it lies outside the common grid; as a value of $\alpha$, it compares across sizes. Each noise draw is seeded by its source, block and grid point alone, so every checkpoint is scored on the same data and noise. The held-out split is part of the data-order release (Appendix~\ref{app:release:order}), so released states can be scored without rebuilding the data.

\textbf{Checkpoints we did not train.} The weights of the S-FLM reference implementation load into our implementation key for key, which lets the harness score released checkpoints of that model (Table~\ref{tab:final}). SFCD's released checkpoint is run through its own code and scored by the same rule.

\section{Architecture, data and training}\label{app:setup}

Table~\ref{tab:shapes} gives the model shapes and Table~\ref{tab:train} the training setup.\label{app:setup:arch}

\subsection{Data}\label{app:setup:data}

Table~\ref{tab:app:corpus} gives the corpus as the build plan sees it. Every token count in this paper is our own, under SmolLM2, and differs from the dataset card's.

\begin{table}[h]
\centering\small
\caption{\textbf{Dolma 3 Mix as tokenized for this work.} Weights are the corpus' own. Epochs are the fraction of each source used in $2$T tokens; no source repeats within $4$T.}
\label{tab:app:corpus}
\setlength{\tabcolsep}{5pt}
\begin{tabular}{lrrr}
\toprule
Source & Weight & Tokens & Epochs at $2$T \\
\midrule
\texttt{common\_crawl}        & $76.07\%$ & $4.7930$T & $0.318$ \\
\texttt{olmocr\_science\_pdfs}& $13.57\%$ & $0.8034$T & $0.338$ \\
\texttt{stack\_edu}           & $6.89\%$  & $0.5210$T & $0.265$ \\
\texttt{finemath\_3plus}      & $2.56\%$  & $0.1794$T & $0.286$ \\
\texttt{rpj\_proofpile\_arxiv}& $0.86\%$  & $0.0546$T & $0.316$ \\
\texttt{dolma1\_7\_wiki\_en}  & $0.04\%$  & $0.0027$T & $0.301$ \\
\midrule
Total & & $6.3540$T & \\
\bottomrule
\end{tabular}
\end{table}

\textbf{One deviation from the corpus' reference recipe.} The web split marks byte ranges for removal by suffix-array deduplication without stating how the removal is to be applied, and we do not apply them; they touch $14.3\%$ of documents and $1.63\%$ of body bytes. Our corpus therefore differs from that of OLMo models of the same lineage, which leaves the comparisons of this paper, all internal, unaffected.

\subsection{Order}\label{app:release:order}

\textbf{What is released.} The order is released as a dataset under ODC-By 1.0: the build plan, which fixes the seed, the mixture weights, the shard layout and the parameters of the keyed permutation; a ledger of the input files with their paths, revision and checksums; the held-out split; and, for every shard, an index of its document boundaries and source rows. The plan can also be rebuilt from its configuration and the pinned corpus revision. Every run reads the order with the same global batch, and a command-line tool expands any step or checkpoint into (source, input file, revision, row) from the plan and the index alone, without rebuilding the data.

\textbf{Two conventions.} Data-parallel ranks interleave over one slot space, so step $t$ consumes exactly the slots $[tB,(t+1)B)$ for global batch $B$; and every shard is tokenized from one input file, so a document's origin is a (file, row) pair.

\textbf{Verification.} Three checks ship with the code. A \emph{round-trip} check compares, for one step, the tokens the dataloader delivered with the documents the tool names, re-read from the source files and re-tokenized. A \emph{rebuild} check regenerates the plan and the held-out split byte for byte. A \emph{held-out} check confirms that no held-out block appears in any training slot. All three pass on a $1$B-token build produced by the same pipeline and can be rerun on the full build. With the released material alone, $170$ sequences from eight training positions, covering all six sources, trace to documents that re-tokenize to exactly the tokens training read.

\subsection{Training}\label{app:setup:train}

\begin{table}[h]
\centering\small
\caption{\textbf{Training setup}, shared by the flow models and the AR twins except where noted. $^{\dagger}$Conventional or inherited defaults that we did not tune.}
\label{tab:train}
\footnotesize\setlength{\tabcolsep}{3pt}
\begin{tabular}{ll}
\toprule
Sequence Length & $2,048$ tokens \\
Global Batch & $4{,}096$ sequences ($8.39$M tokens) \\
Optimizer & AdamW \citep{loshchilov2019adamw} with $\beta=(0.9,0.95)$, $\epsilon=10^{-8}$, wd $0.1^{\dagger}$ \\
Learning Rate & $6{\times}10^{-4}$ (150M), $3{\times}10^{-4}$ (500M, 1B, 3B) with max-norm clip $2.5$ \\
Scheduler & $2$K-step linear warmup$^{\dagger}$, then constant; no run in this paper decays \\
$\alpha$ sampling & antithetic within a rank \\
Parallelism & FSDP2 \citep{zhao2023fsdp,liang2025torchtitan} with up to $32$ nodes $\times\,8$ H200 \\
AC & selective-op activation checkpointing~\citep{chen2016sublinear} with no compilation \\
Checkpoints & every $20$B tokens with optimizer states \\
\bottomrule
\end{tabular}
\end{table}

\textbf{Learning rate.} The learning rates of Table~\ref{tab:train} come from a coarse sweep at the training batch size, meant to exclude wrong orders of magnitude rather than to find an optimum. They follow no scaling rule, one value at 150M and another at every larger size, and should not be interpolated.

\textbf{Window.} The window $[0,\alpha_{\max}]$ of each size (Table~\ref{tab:shapes}) was fixed from an untruncated pilot of $891$M tokens, as the $\alpha$ at which its loss crosses $0.1$ nats, rounded up.

\section{The geometric threshold \texorpdfstring{$\alpha_{\mathrm{geo}}$}{alpha_geo}}\label{app:alphastar}

\textbf{Setup.}
Let $e_i$ be the unit embedding of the target token and $z$ a uniform draw on $S^{d-1}$. The interpolant of Eq.~\ref{eq:slerp} is
$x_\alpha=\big[\sin((1-\alpha)\Theta)\,z+\sin(\alpha\Theta)\,e_i\big]/\sin\Theta$ with $\Theta=\arccos\langle z,e_i\rangle$. In $d$ dimensions $\langle z,e_i\rangle=O(d^{-1/2})$, so $\Theta=\pi/2+O(d^{-1/2})$ and, writing $\theta=\alpha\pi/2$,
\begin{equation}\label{eq:app:slerp}
x_\alpha \;=\; \cos\theta\;z \;+\; \sin\theta\;e_i \;+\; O(d^{-1/2}).
\end{equation}
As $\alpha$ rises the target's component grows and the noise component shrinks; everything below is stated in this limit.

\textbf{Scores.}
The context-free decoder is nearest neighbor in embedding space, $\hat\imath=\arg\max_k\langle x_\alpha,e_k\rangle$. From Eq.~\ref{eq:app:slerp},
\begin{equation}
\langle x_\alpha,e_i\rangle=\cos\theta\,\langle z,e_i\rangle+\sin\theta=\sin\theta+O(d^{-1/2}),
\qquad
\langle x_\alpha,e_j\rangle=\underbrace{\cos\theta\,\langle z,e_j\rangle}_{\text{noise}}+\underbrace{\sin\theta\,\langle e_i,e_j\rangle}_{\text{leakage}} .
\end{equation}
A competitor is helped from two independent directions: noise that happens to point at $e_j$, and the target's own signal leaking into $e_j$ because the two embeddings are similar. The second term is the one an isotropy assumption removes.

\textbf{Bounding the best competitor.}
Put $m=\max_{j}\langle z,e_j\rangle$ and $c=\max_{j\neq i}\langle e_i,e_j\rangle$. Because the maximum of a sum is at most the sum of the maxima,
\begin{equation}\label{eq:app:bound}
\max_{j\neq i}\;\langle x_\alpha,e_j\rangle \;\;\le\;\; B \;:=\; \cos\theta\,m+\sin\theta\,c .
\end{equation}
$B$ is in general \emph{not attained}: $m$ and $c$ are achieved by different tokens, so $B$ describes an adversary that combines the noise luck of one token with the embedding proximity of another. No such token need exist, which is why the resulting threshold is conservative.

\textbf{The threshold.}
Decoding succeeds when the target outscores every competitor. Imposing the strictly stronger requirement $\sin\theta>B$ and chaining with Eq.~\ref{eq:app:bound},
\begin{equation}
\sin\theta \;>\; B \;\ge\; \max_{j\neq i}\langle x_\alpha,e_j\rangle ,
\end{equation}
so success follows. Solving the requirement for $\theta$, and using $\cos\theta>0$ on $(0,\pi/2)$,
\begin{align}
\sin\theta>\cos\theta\,m+\sin\theta\,c
\;&\iff\;
\sin\theta\,(1-c)>\cos\theta\,m
\;\iff\;
\tan\theta>\frac{m}{1-c} \nonumber\\
\;&\iff\;
\alpha>\frac{2}{\pi}\arctan\frac{m}{1-c},
\end{align}
which is Eq.~\ref{eq:alphastar} with $c=c_{\mathrm{mean}}$.

\textbf{How to read $\alpha_{\mathrm{geo}}$.}
The implication runs one way: $\alpha_{\mathrm{geo}}$ \emph{guarantees recovery above itself and is silent below itself}. Since $B$ over-estimates the best competitor, decoding often still succeeds below it, so per-position evidence need not be insufficient there.

\textbf{Its statistical status.} First, $m$ is a maximum over the whole vocabulary rather than over $j\neq i$, because the noise is drawn with no target attached; including $e_i$ raises $m$ only with probability about $1/|V|$, and only in the conservative direction. Second, we evaluate Eq.~\ref{eq:alphastar} at the means of the two maxima, over $1024$ noise draws and $1024$ sampled tokens with a fixed seed, so every size and checkpoint sees the same draws. The maxima are concentrated, with mean and median within $1\%$, so $\alpha_{\mathrm{geo}}$ is a typical-case threshold rather than one holding with a stated probability; using the $95$th percentile of $m$ instead raises it by $15$--$19\%$ at $D=2$T without changing any trend.

\textbf{Relation to the analytic form.}
\citet{hypersphericalflows2026} assume embeddings i.i.d.\ uniform on the hypersphere and bound, by a union bound over the vocabulary, the largest similarity between the noised point and a non-target token, which contains both the noise term and the target's own component, obtaining $\alpha_{\mathrm{geo}}(\delta)=\frac{2}{\pi}\arcsin\big(t_\delta\big)$ with $t_\delta=\sqrt{2\log(2(|V|-1)/\delta)/d}$. At $c_{\mathrm{mean}}=0$, Eq.~\ref{eq:alphastar} recovers this form up to two corrections that do not move during training: it keeps the factor $\cos\theta$ on the noise term, which turns $\arcsin$ into $\arctan$ and changes the threshold by at most $1.1\%$, and it measures $m$ and $c$ separately rather than bounding their combination. Only the factor $1/(1-c_{\mathrm{mean}})$, about $1.1$ for isotropic embeddings and at $20$B, changes during training, and it carries the whole rise of $\alpha_{\mathrm{geo}}$ (Appendix~\ref{app:band:exponents}).

\section{Behind the band}\label{app:band}

\textbf{Why a ratio, and at a matched $D$.} The band's width is a ratio because both boundaries fall with scale and a gap stated in $\alpha$ would not compare across sizes. It also grows with $D$, for the 3B model from $1.22$ at $0.2$T to $3.52$ at $2$T, and the ordering in $N$ of Table~\ref{tab:band} holds only because $D$ is matched there.

\begin{table}[h]
\centering\small
\caption{\textbf{The band early in training and at $D=2$T.} Cross entropy on the common $\alpha$ grid; the two boundaries swap order between the blocks.}
\label{tab:band}
\setlength{\tabcolsep}{5pt}
\begin{tabular}{lrrrrcrrrr}
\toprule
 & \multicolumn{4}{c}{$D=20$B tokens} & & \multicolumn{4}{c}{$D=2$T tokens} \\
\cmidrule(r){2-5}\cmidrule(l){7-10}
 & 150M & 500M & 1B & 3B & & 150M & 500M & 1B & 3B \\
\midrule
held-out CE      & 3.673 & 2.820 & 2.117 & 1.660 & & 2.718 & 1.879 & 1.348 & 0.994 \\
$\alpha_{\mathrm{sol}}$              & 0.121 & 0.096 & 0.076 & 0.063 & & 0.107 & 0.079 & 0.061 & 0.049 \\
$\alpha_{\mathrm{geo}}$ & 0.113 & 0.085 & 0.065 & 0.052 & & 0.257 & 0.242 & 0.202 & 0.174 \\
$c_{\mathrm{mean}}$ & 0.160 & 0.120 & 0.095 & 0.078 & & 0.669 & 0.729 & 0.741 & 0.753 \\
$\alpha_{\mathrm{geo}}/\alpha_{\mathrm{sol}}$ & 0.937 & 0.881 & 0.857 & 0.823 & & 2.417 & 3.051 & 3.294 & 3.523 \\
\bottomrule
\end{tabular}
\end{table}

\subsection{How the geometry scales in \texorpdfstring{$d$}{d}}\label{app:band:exponents}

The noise geometry scales in $d$ as its idealization predicts; the embedding geometry does not. Over the four widths, $m\propto d^{-0.515}$ at $D=2$T and $d^{-0.497}$ at $D=20$B ($R^{2}>0.999$), against the $d^{-1/2}$ expected of a maximum over $|V|$ near-Gaussian projections. $c_{\mathrm{mean}}$, by contrast, rises with size at a fixed $D$ as well as with $D$ (Table~\ref{tab:band}), which the idealization does not allow.

\textbf{The threshold without the anisotropy factor falls.} With $c_{\mathrm{mean}}=0$, Eq.~\ref{eq:alphastar} leaves a threshold driven by $m$ alone, which falls between $D=20$B and $2$T to $0.91$--$0.94$ of its $20$B value, as $m$ drifts down by $6$--$9\%$. The rise of $\alpha_{\mathrm{geo}}$ is therefore carried entirely by the anisotropy factor $1/(1-c_{\mathrm{mean}})$, which grows from $1.08$--$1.19$ at $20$B to $3.0$--$4.0$ at $2$T.

\subsection{The two crossings}\label{app:band:crossings}

At $D=20$B the frontier lies above the threshold at every size, and further above it for larger models, so a larger model starts further behind its own geometry before overtaking it (Table~\ref{tab:app:crossings}).

\begin{table}[h]
\centering\small
\caption{\textbf{The two crossings} (log-$D$ interpolation): $\alpha_{\mathrm{sol}}$ meets $\alpha_{\mathrm{geo}}$, then $\alpha_{\mathrm{geo}}$ meets $\alpha_{\max}$.}
\label{tab:app:crossings}
\setlength{\tabcolsep}{6pt}
\begin{tabular}{lrrrr}
\toprule
 & 150M & 500M & 1B & 3B \\
\midrule
$\alpha_{\mathrm{sol}}/\alpha_{\mathrm{geo}}$ at $D=20$B & 1.07 & 1.14 & 1.17 & 1.22 \\
first crossing, $D$ & $29.2$B & $49.3$B & $62.6$B & $84.8$B \\
first crossing, $\alpha$ & 0.117 & 0.088 & 0.068 & 0.055 \\
second crossing ($\alpha_{\mathrm{geo}}=\alpha_{\max}$), $D$ & $90$B & $218$B & $278$B & $382$B \\
\quad as a fraction of a $2$T budget & $4.5\%$ & $10.9\%$ & $13.9\%$ & $19.1\%$ \\
\bottomrule
\end{tabular}
\end{table}

\subsection{The equivalence point}\label{app:band:alphaeq}

$\alpha_{\mathrm{eq}}$ falls with size both as a value and relative to each size's window (Fig.~\ref{fig:equivalence}, left), so the narrowing window does not account for the trend. Its dependence on $D$ is an order of magnitude weaker: $\mathrm{d}\ln\alpha_{\mathrm{eq}}/\mathrm{d}\ln D$ is $-0.005$, $-0.027$ and $-0.030$ at 150M, 500M and 1B.

\section{Generation, multiple choice and the prefix probe}\label{app:cond}

\subsection{Generation}\label{app:cond:gen}

\textbf{Sampling.} At every $20$B checkpoint to $2$T we draw $256$ samples of $512$ tokens at $64$ function evaluations and temperature $1$, on an $\alpha$ grid inside the model's training window: from pure noise for unconditional generation, and after a clean $64$-token held-out prefix for conditional generation. Both are scored on the $448$ tokens after position $64$.

\textbf{Metrics.} Gen-PPL is scored by SmolLM2-1.7B \citep{smallm}. A sample's rep-4 is the share of its tokens covered by a $4$-gram that occurs in it more than once; a sample collapses when its rep-4 exceeds $0.9$. Held-out text scored at the same positions gives Gen-PPL $9.64$, rep-4 $0.157$ and a collapse rate of $0.39\%$.

\textbf{Smoothing.} The thick lines in Fig.~\ref{fig:layers} are robust LOWESS fits in $\log D$ \citep{cleveland1979robust}.

\textbf{Conditional Gen-PPL inverts.} Over the $400$ checkpoints measured to $2$T, $\log$ conditional Gen-PPL and the collapse rate correlate at $r=-0.73$: the one checkpoint that scores below the real-text floor, 500M at $0.82$T with $5.6$, is the one collapsing most, on $64.8\%$ of its samples. We therefore read conditional Gen-PPL only beside the collapse rate (Table~\ref{tab:midtrain}).

\textbf{The conditional collapse rate.} Every model above 150M passes through a hump midway through training, peaking at $32$--$40\%$ on the smoothed series between $0.6$ and $0.85$T; unconditional generation from the same checkpoints never exceeds $9.4\%$.

\subsection{Multiple choice}\label{app:cond:mcq}

\textbf{The score.} Six tasks \citep{clark2019boolq,mihaylov2018can,bisk2020piqa,zellers2019hellaswag,clark2018think,sakaguchi2021winogrande} are formatted after \citet{gao2024lmeval}; the chance level of their average is $0.375$. For a prompt $x$ and an option $c$ of $|c|$ tokens, let $\mathrm{CE}(c\mid x)$ be the summed cross entropy over the option positions of a sequence in which the prompt is held clean and the option positions are at $\alpha=0$ (pure noise, so the option is read from the prompt alone). Let $\tilde x$ be the prompt with every position replaced by one neutral filler token, a newline. The option scores
\begin{equation}\label{eq:app:mcq}
s(c)\;=\;\frac{1}{|c|}\Big[\mathrm{CE}(c\mid x)\;-\;\mathrm{CE}(c\mid \tilde x)\Big],
\end{equation}
and the option with the smallest score wins. The bracket is a PMI correction \citep{holtzman2021surface}: length, option positions and the noise draw are identical between its two terms, so the difference cancels the per-token term measuring how readily a token is recovered from noise, which otherwise dominates a flow model's option scores. Both normalizations are needed: without the correction the score ranks options by how recoverable their tokens are, and without the division by $|c|$ it ranks them by brevity. Every flow multiple-choice accuracy in this paper uses Eq.~\ref{eq:app:mcq}.

\textbf{The autoregressive twins} are scored by their plain conditional likelihood of each option, with the same length normalization, since neither $\alpha$ nor the PMI correction has a counterpart there. Accuracies are therefore comparable across the two, and scores are not (Table~\ref{tab:final}).

\textbf{Absolute values.} On a quantity pinned at chance a difference counts noise as signal, so we quote absolute values with standard errors. Under uniform $\alpha$ the average stays within $0.354$--$0.388$ at all four sizes over the $27$ checkpoints from $0.5$ to $2$T where it was measured, and the 150M series over twenty checkpoints from $0.1$ to $2$T straddles chance, never more than $3.1\sigma$ from it.

\textbf{What the average contains.} Under uniform $\alpha$ no task shows ability (Table~\ref{tab:position}): HellaSwag, PIQA and WinoGrande stay at chance, OpenBookQA at or below it, and ARC-Easy only slightly above. BoolQ sits below its chance level, at the rate of always answering no. Because the average is a sum of such task-level offsets rather than a quantity continuous in $N$ or $D$, we never draw it as a curve.

\subsection{The prefix probe}\label{app:cond:probe}

\textbf{Construction.} Each item holds the first $64$ tokens of a held-out document clean and offers four $32$-token continuations: the document's own and those of three other documents. A model that uses its context should pick the true continuation almost always, so the probe is built to have a large effect size. The flow models score the continuations at $\alpha=0$ with the same PMI correction as Eq.~\ref{eq:app:mcq} (the continuations share one length, so the length division is moot), and the autoregressive twins by plain log-likelihood. There are $n=1024$ items, chance is $0.25$, and standard errors are $\pm0.011$--$0.016$. Along $D$, the probe was measured at 1B only.

\textbf{The 1B decay.} Under uniform $\alpha$ the 1B model's probe accuracy falls by $4.7\sigma$ from $0.5$T to $2$T (Fig.~\ref{fig:midtrain}), while its cross entropy on the gold continuation improves from $8.34$ to $7.72$ nats. The decay therefore lies in separating the true continuation from the distractors, not in its likelihood under a clean prefix; we leave its cause open.

\section{Clean-prefix mid-training in detail}\label{app:mid}

Table~\ref{tab:position} breaks the multiple-choice results of Table~\ref{tab:midtrain} down by task.

\subsection{Design}\label{app:mid:design}

In expectation $12.5\%$ of a sequence's positions are held clean, neither corrupted nor scored. The mixing probability, the maximum split point and the length of $100$B tokens were fixed before any branch was run. Each branch resumes the uniform run's optimizer, schedule and dataloader position, and $k$ depends on the sequence's slot index alone. The 150M--1B branches at $1.9$T also resume the random-number state, so the two runs see identical $\alpha$ and noise draws; the 3B branch at $1.9$T and the two early 1B branches start from states without a kept generator state and draw a different, identically distributed stream.

\subsection{Results at a matched budget}\label{app:mid:main}\label{app:mid:cost}\label{app:mid:both}

Per task (Table~\ref{tab:position}), BoolQ supplies most of the multiple-choice gain at every size, and ARC-Easy adds to it at 1B and 3B. On the probe the autoregressive twins score $0.797$, $0.852$ and $0.867$ at 150M, 500M and 1B. Under mid-training, unconditional Gen-PPL at $2$T stays within its checkpoint-to-checkpoint spread and the unconditional collapse rate below $2\%$. Its cost in held-out cross entropy is in Table~\ref{tab:midtrain}, and what instruction tuning adds on top in Table~\ref{tab:app:dolci}.

\begin{table}[t]
\centering\small
\caption{\textbf{Multiple choice by task at $D=2$T}, under uniform $\alpha$ and after clean-prefix mid-training. First row: chance. Standard errors are $.004$--$.022$ per task and $.005$ for the average. Earlier releases, the final model and the twins are in Table~\ref{tab:final}.}
\label{tab:position}
\setlength{\tabcolsep}{5pt}
\begin{tabular}{lccccccc}
\toprule
 & BoolQ & OBQA & PIQA & HSwag & ARC-E & WinoG & Avg. \\
\midrule
\emph{Chance} & \emph{.500} & \emph{.250} & \emph{.500} & \emph{.250} & \emph{.250} & \emph{.500} & \emph{.375} \\
\midrule
\greyrow \multicolumn{8}{c}{\textbf{Uniform $\alpha$}} \\
150M & .377 & .214 & .513 & .251 & .332 & .486 & $.362$ \\
500M & .379 & .196 & .504 & .253 & .314 & .481 & $.354$ \\
1B   & .386 & .220 & .526 & .258 & .266 & .485 & $.357$ \\
3B   & .379 & .252 & .520 & .247 & .300 & .517 & $.369$ \\
\greyrow \multicolumn{8}{c}{\textbf{Clean-prefix mid-training}} \\
150M & .621 & .218 & .521 & .263 & .303 & .493 & $.403$ \\
500M & .611 & .256 & .541 & .282 & .356 & .470 & $.419$ \\
1B   & .611 & .232 & .573 & .286 & .385 & .518 & $.434$ \\
3B   & .617 & .226 & .552 & .294 & .409 & .492 & $.432$ \\
\bottomrule
\end{tabular}
\end{table}

\subsection{The budget axis}\label{app:mid:daxis}

The 1B branches leave the uniform run at $0.4$, $0.9$ and $1.9$T. At all three landing points, SFT-format cross entropy is lower with the prompt clean than corrupted after mid-training, and more than twice as high under uniform $\alpha$. The multiple-choice gain along $D$ is carried by BoolQ; the other five tasks move by less than $2\sigma$.

\subsection{Weight averaging}\label{app:mid:ema}

The averaged 3B model is the exponential moving average of \citet{meterez2026anytime}, $\bar\theta_{t+1}=\tau_t\bar\theta_t+(1-\tau_t)\theta_t$ with $\tau_t=2^{-f/t}$ at step $t$ and $f=100$. We reconstruct it after the fact from the states saved every $20$B tokens: between consecutive states at steps $T_{i-1}$ and $T_i$ the average decays by $(T_{i-1}/T_i)^{f\ln 2}$, and the weight of that interval falls on the state at $T_i$.

\section{Post-training in detail}\label{app:post}

\subsection{General-purpose post-training}\label{app:post:dolci}

\textbf{Recipe.} One epoch of Dolci-Instruct-SFT is $1{,}107$ steps and $1.16$B tokens, with examples packed by concatenating and chunking. The prompt is held clean and unscored and the response is corrupted at the sequence's own $\alpha$, which makes the recipe the supervised counterpart \citep{wei2022flan,ouyang2022instructgpt} of the mid-training mask of \S\ref{sec:recover:design}. The same recipe is applied to the mid-trained checkpoints, to the uniform-$\alpha$ checkpoints and to the autoregressive twins.

\textbf{Results} are in Table~\ref{tab:app:dolci}, with the cost in held-out cross entropy measured on the common grid.

\begin{table}[h]
\centering\small
\caption{\textbf{One epoch of Dolci-Instruct-SFT from the $2$T bases.} Multiple-choice accuracy before and after, and the rise in held-out cross entropy. Standard errors are about $.005$.}
\label{tab:app:dolci}
\setlength{\tabcolsep}{5pt}
\begin{tabular}{llcccc}
\toprule
Base & & 150M & 500M & 1B & 3B \\
\midrule
Uniform $\alpha$ & MCQ before & .362 & .354 & .357 & .369 \\
 & MCQ after & .398 & .422 & .407 & .436 \\
 & CE cost & $+7.7\%$ & $+8.5\%$ & $+8.1\%$ & $+11.3\%$ \\
\midrule
Clean-prefix mid-trained & MCQ before & .403 & .419 & .434 & .432 \\
 & MCQ after & .417 & .429 & .440 & .446 \\
 & CE cost & $+3.8\%$ & $+3.0\%$ & $+2.4\%$ & $+5.4\%$ \\
\midrule
Autoregressive & MCQ before & .481 & .551 & .604 & --- \\
 & MCQ after & .446 & .547 & .614 & --- \\
\bottomrule
\end{tabular}
\end{table}

\subsection{Task-specific setup}\label{app:post:setup}

\textbf{Recipe.} Each TinyGSM \citep{liu2023tinygsm} example fills one $512$-token block, and the loss is taken on the padding as well as on the answer; the prompt is clean and unscored as in Appendix~\ref{app:post:dolci}. The learning rate is $3\times10^{-4}$ in every cell, and checkpoints are read at $20$, $40$ and $65$B finetuning tokens. ``None'' is a randomly initialized model of the same size given the same finetuning. The ten conditions are the four sizes from none, the four sizes from their $2$T mid-trained checkpoints, and the 1B model from its $0.5$T and $1$T mid-trained branches (\S\ref{sec:recover:daxis}).

\textbf{Layout.} Examples are not packed, as in the reference implementation's TinyGSM scripts: under packing, the positions after an answer learn to start a new problem, which a flow model, decoding the whole block at once, then writes in competition with the answer.

\textbf{Scoring and sampling.} GSM8K \citep[$n=1319$;][]{cobbe2021training} is prompted $0$-shot, and the generated Python is executed; a generation is correct when the program's returned value matches the gold answer \citep{gao2023pal,chen2021codex}. Generation uses $64$ function evaluations at temperature $1.0$ with a greedy final decode. The \emph{exact} sampler marginalizes the per-step velocity over the vocabulary and is the reference implementation's default; the \emph{top-1} sampler aims the velocity at the arg-max token. Beside accuracy we report the executable rate, the fraction of generations whose function returns without an exception or a timeout.

\subsection{Results}\label{app:post:axes}\label{app:post:levels}\label{app:post:anomaly}

Accuracy rises from $20$ to $40$ to $65$B finetuning tokens in every condition, with both samplers. Table~\ref{tab:app:tinygsm} gives the end points of Fig.~\ref{fig:tinygsm}.

\begin{table}[h]
\centering\small
\caption{\textbf{GSM8K after $65$B tokens of TinyGSM finetuning}, $0$-shot, scored by execution ($n=1319$); executable rate in parentheses. ``None'': random initialization; other bases are mid-trained.}
\label{tab:app:tinygsm}
\setlength{\tabcolsep}{5pt}
\begin{tabular}{llcccc}
\toprule
Sampler & Base & 150M & 500M & 1B & 3B \\
\midrule
exact & none & 0.0197 (0.469) & 0.0637 (0.522) & 0.0743 (0.513) & 0.0879 (0.484) \\
      & $2$T & 0.0576 (0.588) & 0.1168 (0.625) & 0.1251 (0.644) & 0.1266 (0.655) \\
\midrule
top-1 & none & 0.0485 (0.515) & 0.1334 (0.594) & 0.1471 (0.577) & 0.1509 (0.598) \\
      & $2$T & 0.1137 (0.652) & 0.2138 (0.754) & 0.2237 (0.797) & 0.2055 (0.741) \\
\bottomrule
\end{tabular}
\end{table}

\end{document}